\documentclass[letterpaper, 10 pt, conference]{ieeeconf}  

\IEEEoverridecommandlockouts                              

\usepackage{graphicx}
\usepackage[ruled]{algorithm2e}
\usepackage{siunitx}
\usepackage{algpseudocode}
\usepackage{amsmath} 
\graphicspath{{figures//}}
\makeatletter
\let\NAT@parse\undefined
\makeatother
\usepackage[colorlinks,allcolors=green]{hyperref}
\usepackage[noadjust]{cite}

\title{\LARGE \bf
FANTrack: 3D Multi-Object Tracking with Feature Association Network
}

\author{Erkan Baser$^{1}$, Venkateshwaran Balasubramanian$^{2}$ \textsuperscript{*}, Prarthana Bhattacharyya$^{3}$ \textsuperscript{*}, Krzysztof Czarnecki$^{3}$
\thanks{$^{1}$ Erkan Baser was affiliated to Waterloo Intelligent Systems Engineering Lab, University of Waterloo, 200 University Ave W, Waterloo, ON N2L 3G1. {\tt\small erkanbaser@gmail.com}}
\thanks{$^{2}$ Venkateshwaran Balasubramanian is with the David R. Cheriton School of Computer Science, University of Waterloo, 200 University Ave W, Waterloo, ON N2L 3G1. {\tt\small venk.b@outlook.com}}
\thanks{$^{3}$ Prarthana Bhattacharyya and Krzysztof Czarnecki are with the Department of Electrical and Computer Engineering, University of Waterloo, 200 University Ave W, Waterloo, ON N2L 3G1.}
\thanks{{\tt\small \{p6bhatta,k2czarne\}@uwaterloo.ca}}
\thanks{\textsuperscript{*} denotes equal contribution}
\thanks{\newline \textcopyright \enspace 2019 IEEE. Personal use of this material is permitted. Permission from IEEE must be obtained for all other uses, in any current or future media, including reprinting/republishing this material for advertising or promotional purposes, creating new collective works, for resale or redistribution to servers or lists, or reuse of any copyrighted component of this work in other works.}
}

\begin{document}
\maketitle
\pagestyle{empty}

\begin{abstract}
We propose a data-driven approach to online multi-object tracking (MOT) that uses a convolutional neural network (CNN) for data association in a tracking-by-detection framework. The problem of multi-target tracking aims to assign noisy detections to a-priori unknown and time-varying number of tracked objects across a sequence of frames. A majority of the existing solutions focus on either tediously designing cost functions or formulating the task of data association as a complex optimization problem that can be solved effectively. Instead, we exploit the power of deep learning to formulate the data association problem as inference in a CNN. To this end, we propose to learn a similarity function that combines cues from both image and spatial features of objects. Our solution learns to perform global assignments in 3D purely from data, handles noisy detections and a varying number of targets, and is easy to train. We evaluate our approach on the challenging KITTI dataset and show competitive results. Our code is available at \url{https://git.uwaterloo.ca/wise-lab/fantrack}.
\end{abstract}

\section{INTRODUCTION}
Multi-object tracking (MOT) is a critical problem in computer vision and has received great attention due to its widespread use in applications such as autonomous driving, robot navigation, and activity recognition. It is the problem of finding the optimal set of trajectories of objects of interest over a sequence of consecutive frames. Most of the successful computer vision approaches to MOT have focused on the \textit{tracking-by-detection} principle \cite{Okuma2004ABP, trackdet}. This paradigm allows the problem to be divided into two steps. First, an object detector is used to identify the potential locations of objects in the form of bounding boxes, and then a discrete combinatorial problem is solved to link these noisy detections over time to form trajectories. Despite decades of research, the status quo of tracking is far from reaching human accuracy. Current challenges to the problem include a varying and a-priori unknown number of targets; incorrect and missing detections; changing appearances of targets due to sensor motion, illumination, and angle of view; frequent occlusions, and abrupt changes in motion.
\begin{figure}[t]
\begin{center}
\includegraphics[width = 7cm, height = 8cm]{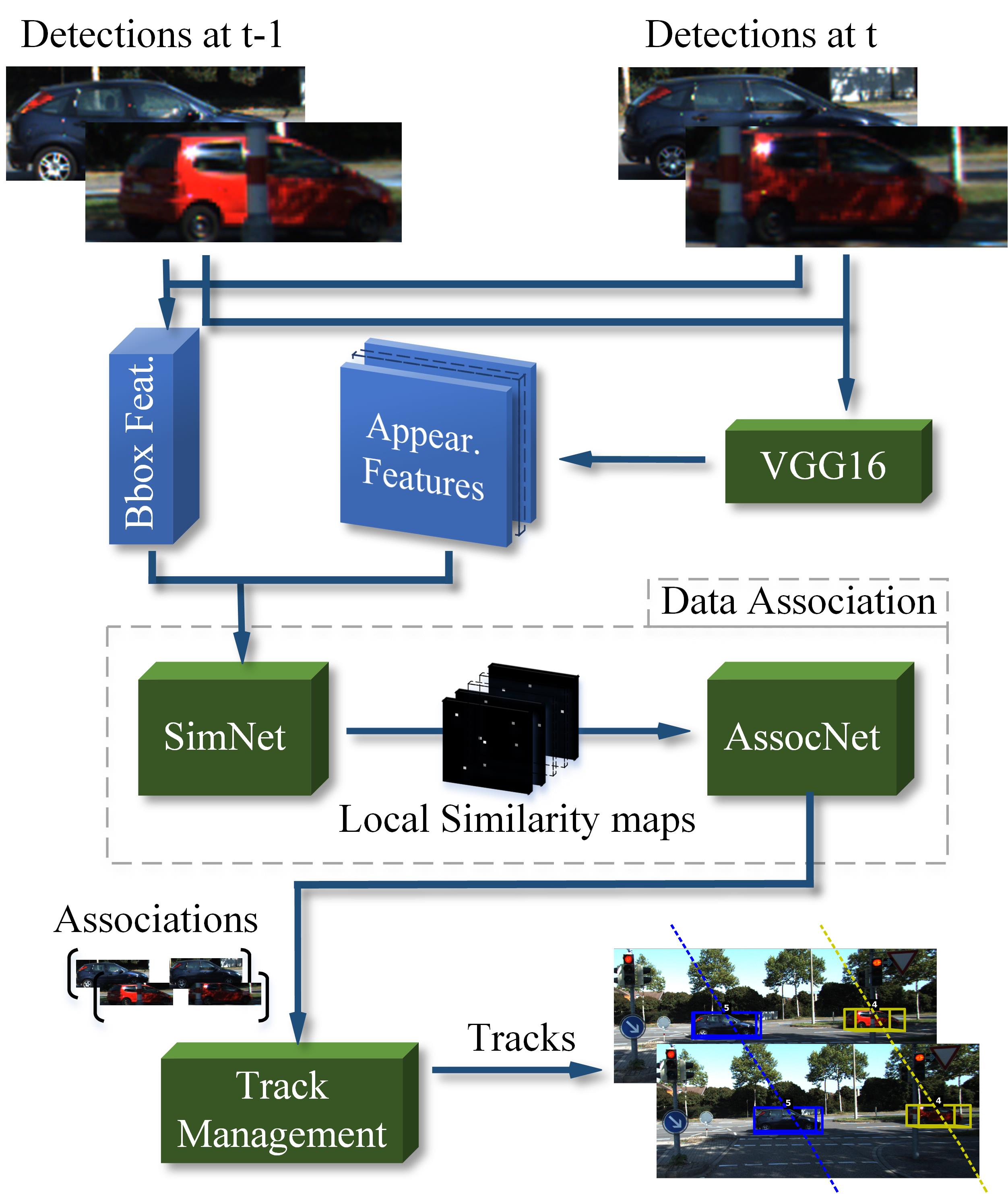}
\end{center}
\caption{Overall architecture of the proposed approach}
\label{end_to_end}
\end{figure}

\par  The linking step called \textit{data association} is arguably the most difficult component of MOT. Traditional \textit{batch} methods usually formulate MOT as a \textit{global} optimization problem, with the assumption that detections from all future frames are available, and solve it by mapping it to a graph based min-cost flow algorithm \cite{netflow, globalopt}. \textit{Online} Markovian formulations of MOT on the other hand often employ greedy or bipartite graph matching methods like the Hungarian algorithm to solve the assignment problem \cite{Munkres57algorithmsfor, hungarian, breitenstein2009RobustTU}. \textit{Online} approaches are well suited to real-time applications such as tracking road-traffic participants. The success of the final associations is also dependent on the similarity functions used to match the targets and detections. Traditionally cost functions have been handcrafted with representations based on color histograms, bounding box position, and linear motion models \cite{appearance, pos}, but have failed to generalize across tasks and for complex tracking scenarios. Recently, deep neural network architectures have shown superior performance in many vision based tasks. Milan et al. proposed the first end-to-end formulation for MOT, using a recurrent neural network (RNN) to solve the assignment problem for each target independently based on Euclidean cost \cite{milan}. However, the use of convolutional neural networks (CNNs), which are easier to train than RNNs, in order to solve the association problem while also learning the cost function has not yet been investigated.

\par In this paper, we propose an \textit{online} MOT formulation that casts the assignment problem as inference in a CNN. We present a two-step learning based approach (see Fig. \ref{end_to_end}). The first step learns a similarity function that takes advantage of both visual and 3D bounding box data to yield robust matching costs. The second step trains a CNN to predict discrete target assignments from the computed pair-wise similarities. The benefit of our proposal is that it is easy to train, takes care of a varying number of targets and noisy detections, and provides a simple way to consider all the targets while making associations. We empirically demonstrate on the KITTI tracking dataset \cite{Kitti_Dataset} that: (i) Our approach can solve the multi-target association problem by performing inference using CNNs. (ii) It can integrate image based appearance and 3D bounding box features to get a discriminative as well as generalized feature representation, thereby learning a robust cost function for association. (iii) We show competitive qualitative and quantitative 3D tracking results compared to the state of the art. 


\section{Literature Review}
\subsection{Data Association in MOT}
Classical approaches solve the data association problem by considering multiple hypotheses for an assignment (MHT) \cite{mht}, or by jointly considering all possible assignment hypotheses (JPDA) \cite{jpda}. These formulations prove to be very computationally intensive, however. 
\par Many recent works process sequences in \textit{batch} mode, using a graph-based representation with detections as nodes and possible assignments as edges. The optimization is then cast as a linear program solved to (near) global optimality with relaxation, min-cost or shortest path algorithms \cite{relax, flow, path}. More complex optimization schemes include MCMC \cite{mcmc} and discrete-continuous settings \cite{andrienko}. However, \textit{global} optimization formulations are unsuited to real-time applications like autonomous navigation. 
\par \textit{Online} methods estimate the current state using the information only from the past frames and the current one. Commonly used state-estimators include the Kalman filter \cite{kalman} for linear motion and particle filters \cite{particle} for multi-modal posteriors. The two-frame association problem is often solved using a greedy or Hungarian algorithm \cite{hungarian}. Approaches based on local associations tend to be susceptible to track fragmentation and noisy detections, however.

\par \textit{Deep learning} has achieved state-of-the-art results in perception tasks like image classification, segmentation, and single object tracking. Milan et al. proposed the first fully end-to-end multi-object tracking method based on deep learning. The method predicts the assignment of each target, one at a time, using an RNN~\cite{milan}. In contrast, our approach feeds all detections and their learned similarity scores at once into a CNN to predict the assignments. Our model is easier to optimize than an RNN, handles noisy detections and a varying number of targets, and considers all targets at once when performing assignments.

\subsection{Measuring Similarity}
Tracking algorithms have used distance functions such as Euclidean \cite{euclidean} and Mahalanobis distance \cite{mahalanobis} as matching costs for data association. Other similarity measures include color-based appearance features \cite{app2}, SIFT-like features \cite{sift}, and linear and non-linear motion models and their various weighted combinations~\cite{afld}. These tediously hand-crafted features fail to generalize across complex scenarios and backgrounds, however.
\par Recent works explore learning pairwise costs using deep structured SVM \cite{netflow}, CNNs \cite{quadcnn}, and RNNs \cite{amir}. For CNNs, similarity learning often exploits Siamese networks. Leal-Tai\^xe et al. \cite{leal-taixe} and Frossard et al. \cite{frossard} use them to learn descriptors for matching with multi-modal inputs. While we also use Siamese networks to learn \textit{generalized} and \textit{discriminative} features from 3D object configurations and visual information conditioned on similarity, we adapt our objective function to use the cosine-similarity metric with hard-mining which has a positive impact on convergence. 

\begin{figure*}[t]
\vspace{1.5em}
\begin{center}
\includegraphics[width=12cm, height=4cm]{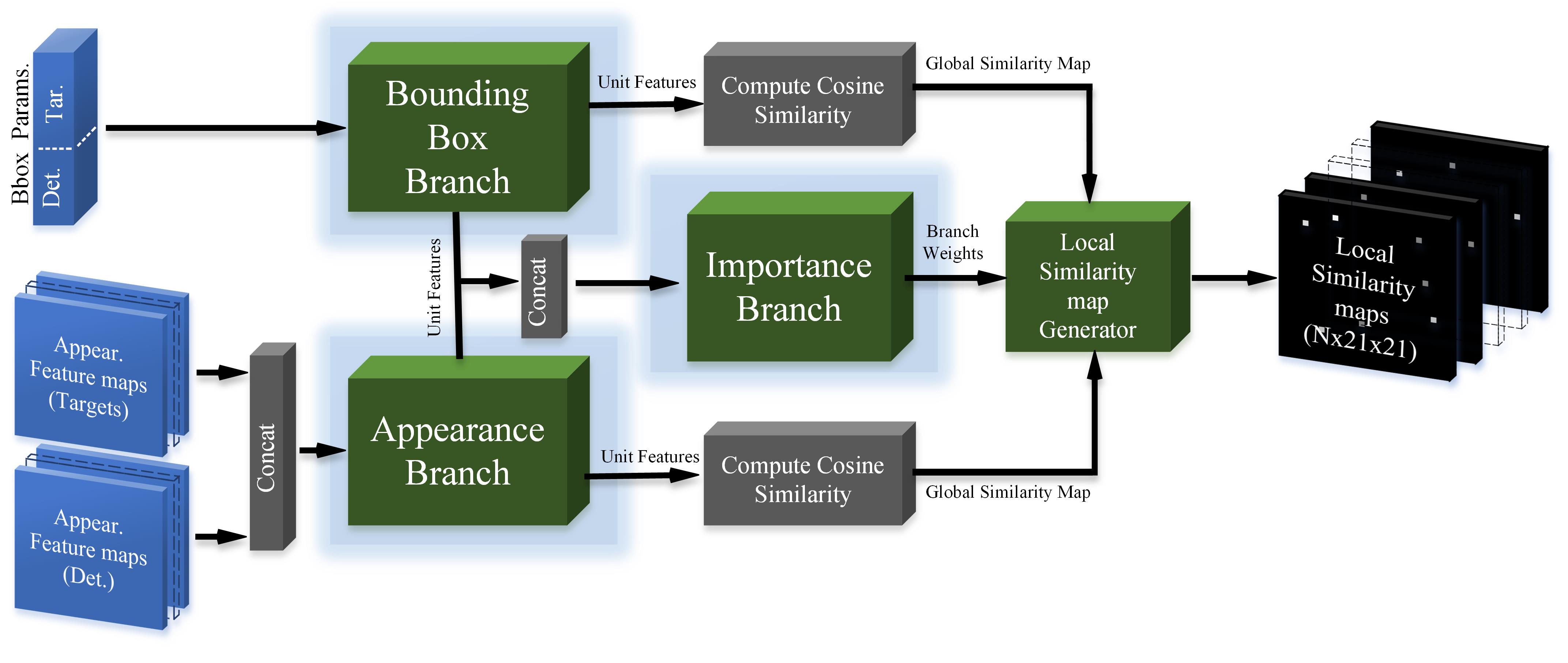}
\end{center}
\caption{Architecture of the proposed Siamese network for similarity learning. The branches highlighted in blue have trainable parameters.}
\label{fig_simNet}
\vspace{-1.5em}
\end{figure*}

\section{Our Approach}
Our proposed framework is based on tracking by detection paradigm. Our problem setup assumes at any time instant $t$ we have $N$ number of targets, $M$ number of detections and track labels for every $i^{th}$ track. We use AVOD \cite{AVOD} as our 3D object detector since it achieves state-of-the-art results on KITTI and is open-source, but in principle, any other 3D object detector could be used. The motivation for building FANTrack is to leverage the power of Siamese networks to model the similarities between targets and detections, CNNs to solve the data association problem in MOT, and an online track management module to update, initialize and prune tracks. We describe these modules in the following sections. 

\subsection{Similarity Network}
Figure~\ref{fig_simNet} gives an overview of our proposed similarity network \textit{SimNet}. The network has two \textit{input} pairs with each pair corresponding to target and detection data, and consists of 3D bounding box parameters ($1\times7$ dimensional vector) and image convolutional features ($7 \times 7 \times 320$ dimensional vector). The \textit{output} from \textit{SimNet} is a set of $N_{max}$ number of maps for each existing target corresponding to a \textit{local} $5 \si{\m} \times 5 \si{\m}$ region around the target and of $0.5 \si{\m}$ resolution. These output maps contain the similarity scores in each target's local neighbourhood with respect to all detections at a particular time step. The \textit{SimNet} output is subsequently used for data association. 
\par \textit{Functionally}, \textit{SimNet} computes a similarity score for every detection and target pair. It has two branches: a \textit{bounding box branch} and an \textit{appearance branch}, each of which uses a trainable Siamese network to learn object representations conditioned on whether two objects are similar or not. The outputs of these branches are vector representations of targets and detections. Their respective contribution towards the final similarity score computation is weighted using the \textit{importance branch}. Finally, cosine-similarities of each target-detection vector representation are computed and the scalars are mapped to their corresponding positions on the above-mentioned set of local maps. 
\par We describe our formulation of \textit{SimNet} in the remainder of this sub-section.

\subsubsection{Bounding Box Branch}
The bounding box branch outputs a discriminative, robust vector representation for 3D object configurations of targets and detections, conditioned on whether they are similar or not. We train a Siamese network with \textit{stacked} input pairs of target and detection 3D bounding boxes for this purpose. The 3D bounding boxes are defined by their centroids $\left(x, y, z \right)$, axis-aligned dimensions $\left( l, w, h \right)$, and rotation around the $z$-axis $\left( \theta_z \right)$  in the ego-car's IMU/GPS coordinates. To prevent learning variations induced due to ego-motion, detection centroids are converted to coordinates at a common time-step using GPS data. 

\begin{figure}[t]
\begin{center}
\includegraphics[width=0.95\linewidth]{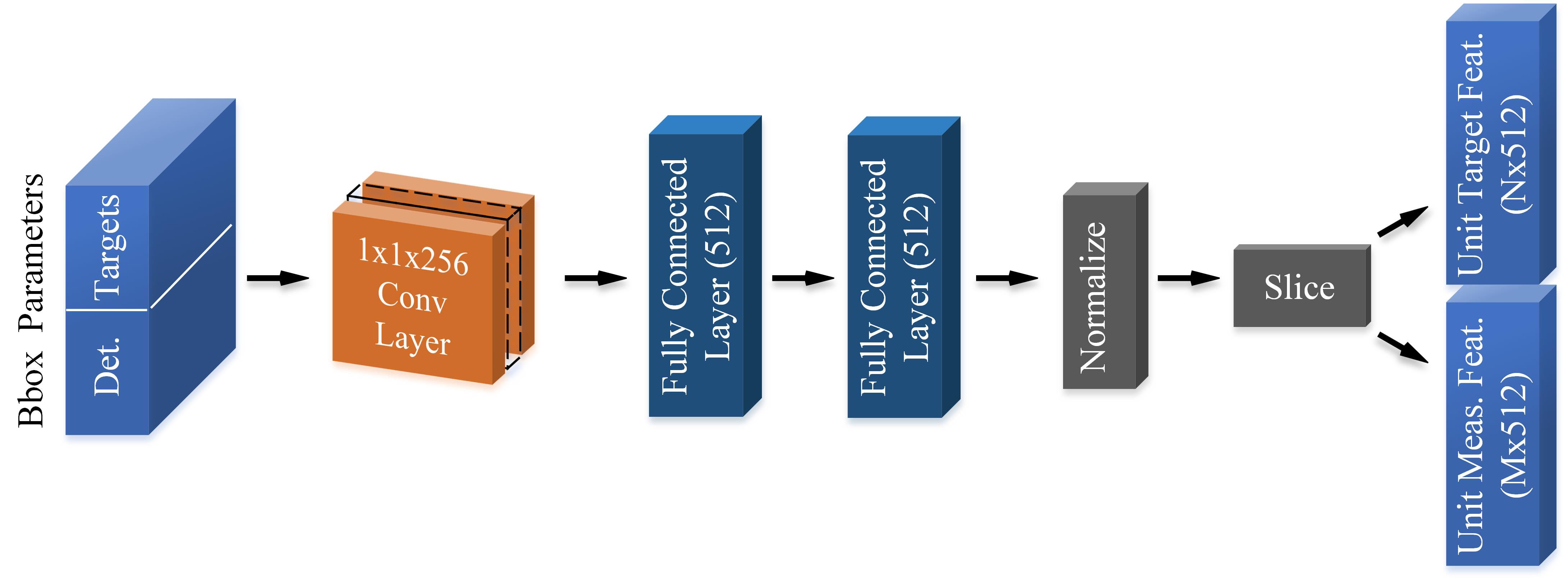}
\end{center}
\caption{Detailed architecture of the bounding box branch. Inputs are the concatenated bounding boxes of targets and detections. Outputs are sliced unit feature vectors.}
\label{fig_bbox_branch}
\end{figure}

\textbf{\textit{Architecture:}} The input to this branch is a $(N+M)\times 1 \times 7$ tensor where the third dimension consists of the $7$ bounding box parameters defined above. The inputs are fed to a convolutional layer with $256$ $1 \times 1$ filters to capture complex interactions across the $7$ channels by pooling the parameters of each targets and detections independently \cite{Net_in_Net}. These object-independent features are then fed into two fully-connected layers with $512$ neurons, with dropout regularization. We apply L2 normalization on the output features, and henceforth refer to the result as \textit{unit} features. Finally, the unit features of dimensions $(N+M)\times512$ are sliced along the first dimension into target features and detection features using their respective counts (see Fig \ref{fig_bbox_branch}). These are used to compute the bounding box cosine similarities as described in subsection \textit{A.4}. We use batch normalization and leaky-ReLU across all layers.

\subsubsection{Appearance Branch}
The appearance branch outputs a robust and invariant vector representation for 2D visual cues of targets and detections conditioned on whether they belong to similar or dissimilar objects. We train another stacked Siamese network for this purpose. As its input, we concatenate convolutional features of targets and detections obtained from AVOD's (\cite{AVOD}) image feature extractors. Specifically, we use the second layer's convolutional features and the interpolated fourth layer's convolutional features. This is because the low-level features are local and more discriminative whereas high-level features are abstract, and are more invariant to appearance changes \cite{CRBM, Visual_conv}.

\begin{figure}[ht]
\vspace{1.5em}
\begin{center}
\includegraphics[width=0.95\linewidth]{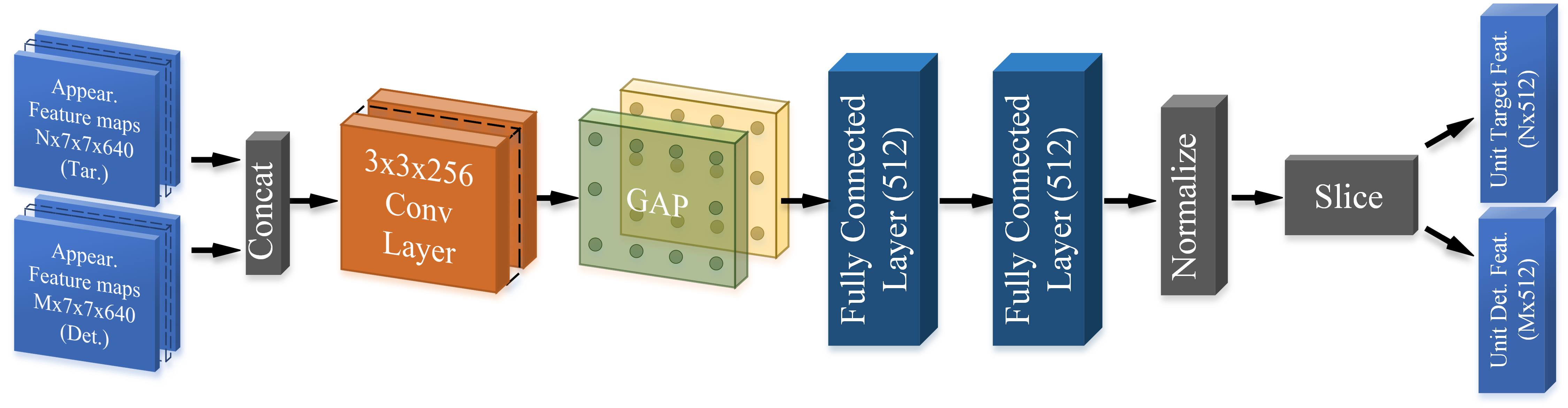}
\end{center}
\caption{Detailed architecture of the appearance branch. Inputs are concatenated appearance feature maps of targets and detections. Outputs are sliced unit feature vectors.}
\label{fig_appear_branch}
\vspace{-1.5em}
\end{figure}

 \textbf{\textit{Architecture:}} The input to this branch is a $(N+M) \times 7 \times 7 \times 320$ convolutional feature. The architecture of the branch is shown in Fig.~\ref{fig_appear_branch}. First, we apply 256 $3 \times 3$ convolutions to obtain promising features for similarity learning by preserving the spatial size of the input. Before flattening the feature maps for the fully-connected layers with $512$ neurons, the Global Average Pooling (GAP) \cite{lin2013network} layer extracts one abstract feature from each feature map. Similar to the bounding box branch, L2 normalization yields a vector of dimension $(N+M)\times512$. As in the case of the bounding box branch, the $(N+M)\times512$ features are sliced along the \textit{first} dimension to obtain appearance features of detections and targets to compute the appearance cosine similarities.
\subsubsection{Importance Branch}
\begin{figure}[t]
\vspace{1.5em}
\begin{center}
\includegraphics[width=0.95\linewidth]{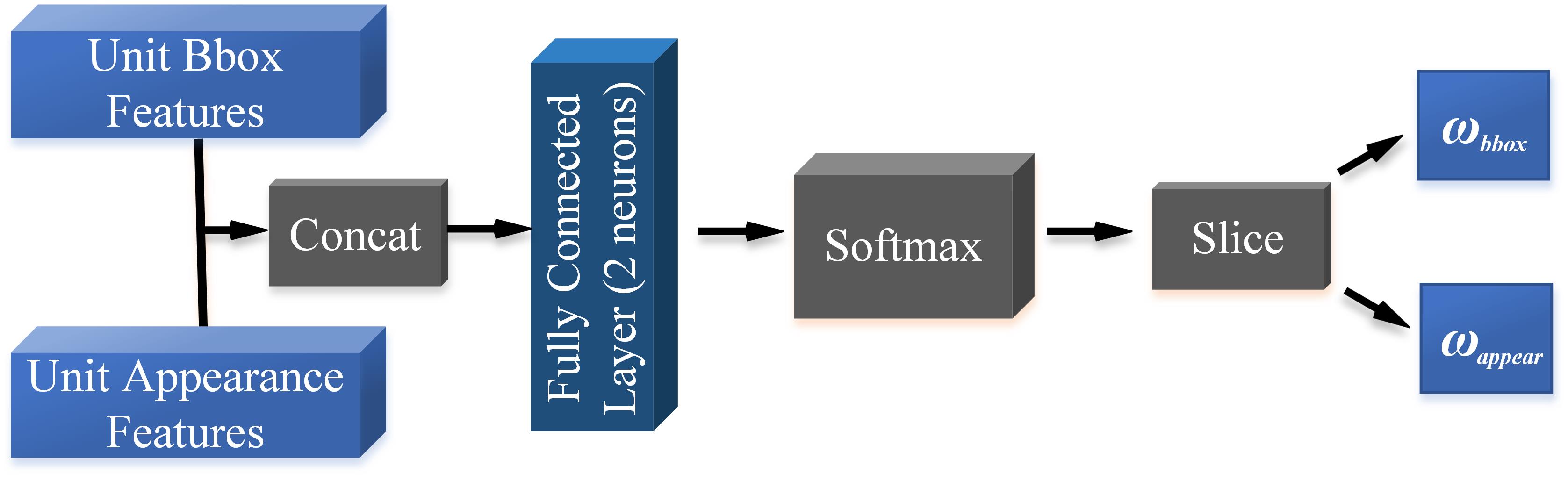}
\end{center}
\caption{Detailed architecture of the importance branch. The inputs are the unit bounding box and appearance features of both targets and detections. Outputs are branch weights for bounding box and appearance branches. These weights can be further sliced for targets and detections separately.}
\label{fig_importance_branch}
\vspace{-1.5em}
\end{figure}

The aim of this branch is to determine the \textit{relative importance} of the bounding box and appearance features in the computation of the final cosine similarity score (see Fig.~\ref{fig_importance_branch}). 
\\\textbf{\textit{Architecture:}} The inputs to this branch are the unit features from the other two branches.  First, the vector representation of an object obtained from the appearance and bounding box branches is concatenated to form a single vector (dimension $1024$). Then, a fully-connected layer with two neurons, ReLU activation, and a softmax layer that computes two scalars indicating importance weights (probabilities) of the two branches, for each target and detection. The importance weights obtained ($\omega _{bbox}$ and $\omega _{appear}$) are normalized to sum up to unity.

\subsubsection{Similarity Maps}
\begin{figure}[t]
\vspace{1.5em}
\begin{center}
\includegraphics[width=1\linewidth]{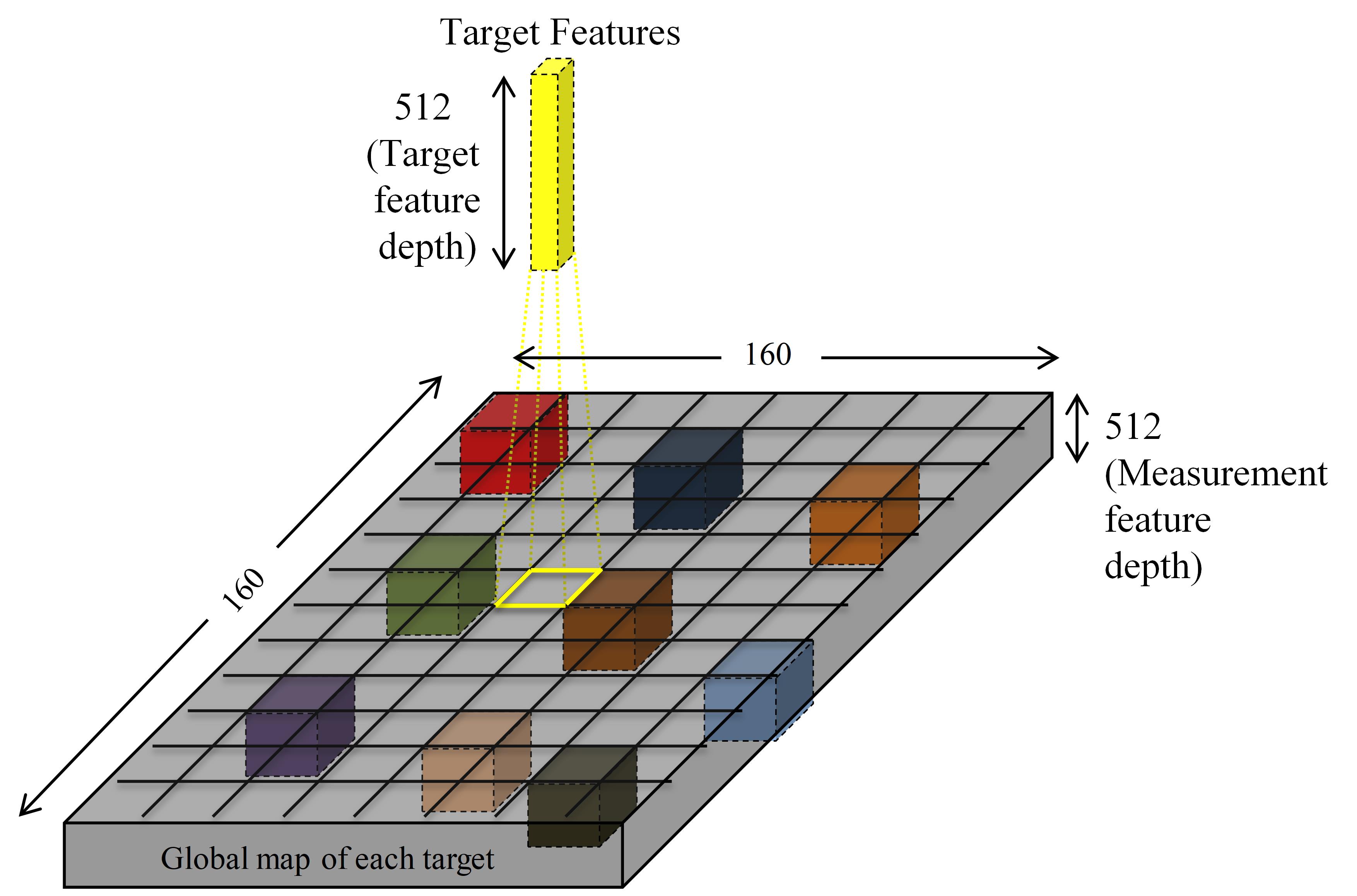}
\end{center}
\caption{Construction of global similarity map, one for each target. The target feature vector (yellow solid) is convolved with those of the detections to compute the similarity scores. Locations that do not include a detection feature vector are filled by zero vectors.}
\label{global_map_construction}
\vspace{-1.5em}
\end{figure}

 A similarity map (see Fig.~\ref{global_map_construction}) is computed for every target (for $N$ targets we have $N$ similarity maps) and contains its cosine similarity scores with all the detections within a 2D region of interest ($\left[ -40,40 \right] \times \left [ 0,80\right]$ \si{\m}) in the ego-car's IMU/GPS coordinates. This map is referred to as the \textit{global} similarity map. To compute the similarity scores, we perform the following:
 \\ $\textbf{\textit{i}})$ Each global map is split into grids with 0.5 \si{\m} resolution.
 \\ $\textbf{\textit{ii}})$ The detection appearance and bounding box features are positioned into the appropriate grid locations based on their location of detection.
 \\ $\textbf{\textit{iii}})$ The target appearance and bounding box features are used as kernels to compute the similarity scores by the convolution with strides equal to $1$. The computed scores correspond to cosine similarities as the features are normalized to unit vectors by the network branches. 
 \par Local similarity maps are then obtained from the global similarity maps for each target by cropping them around the target's local $5\si{\m}\times5\si{\m}$ region corresponding to $10\times10$ cells. \textit{SimNet} thus finally outputs $N\times21\times21$ local similarity maps to be used for data association.
 
\subsubsection{Loss Function} \label{training_siamese}
To learn the trainable parameters of the appearance branch, bounding box branch, and importance branch, we use the \textit{weighted cosine distance} given by:
\begin{flalign}{\label{eq1}}
L(\Theta_1) = \frac{1}{N^+} \sum\nolimits_{i=1}^N{w_{skew}^{(i)}} \times 
w_{cost}^{(i)}  \times  \nonumber \\ \left( 1 - y^{\left(i\right)} \times \hat{y}^{\left(i\right)} \left(\Theta_1\right) \right)
\end{flalign}
\noindent
where ${\Theta _1 }$ is the network parameters, ${N^ + }$ is the number of examples with nonzero weights, $y^{\left( i \right)}$ denotes the ground truth value of  the $i^{th}$ example, i.e., $y^{(i)} \in \left\{ { - 1,1} \right\}$. ${\hat y^{\left( i \right)} }$ is the estimated cosine similarity score computed using the cosine similarities from the two branches and their normalized importance weights as follows:
\begin{flalign}{\label{eq2}}
\hat y^{\left( i \right)}\left( {\Theta _1 } \right)  = \omega _{bbox}\left( {\Theta _1 } \right)^{\left( i \right)}  \times \hat y_{bbox}^{\left( i \right)}\left( {\Theta _1 } \right)  + \nonumber \\ \omega _{appear}^{\left( i \right)}\left( {\Theta _1 } \right)  \times \hat y_{appear}^{\left( i \right)}\left( {\Theta _1 } \right)
\end{flalign}
\noindent
$w_{skew}^{\left( i \right)}$ is the weight used to remove the imbalance of negative examples in the training dataset. $w_{cost}^{\left( i \right)}$ scales the loss function according to how easy or hard it is to distinguish between each pair of examples so that the training can revolve around a sparse set of the selected hard examples \cite{Focal_loss}.

\subsubsection{Creating training examples for \textit{SimNet}} In this section, we describe creating positive and negative pairs of examples by augmentation from the KITTI training set to train the similarity network. A new bounding box proposal is a positive pair if its intersection over union (IoU) with its ground truth on images exceeds $0.8$. The selected IoU thresholds should be at least greater than those used for the final detections in AVOD $(0.65)$ \cite{AVOD}. In addition, the diversity among bounding box proposals for each object is maintained by rejecting a new proposal whose IoUs with existing ones are greater than $0.95$.


\begin{figure*}[t]
\vspace{1.5em}
\begin{center}
\includegraphics[width=12cm, height=4cm]{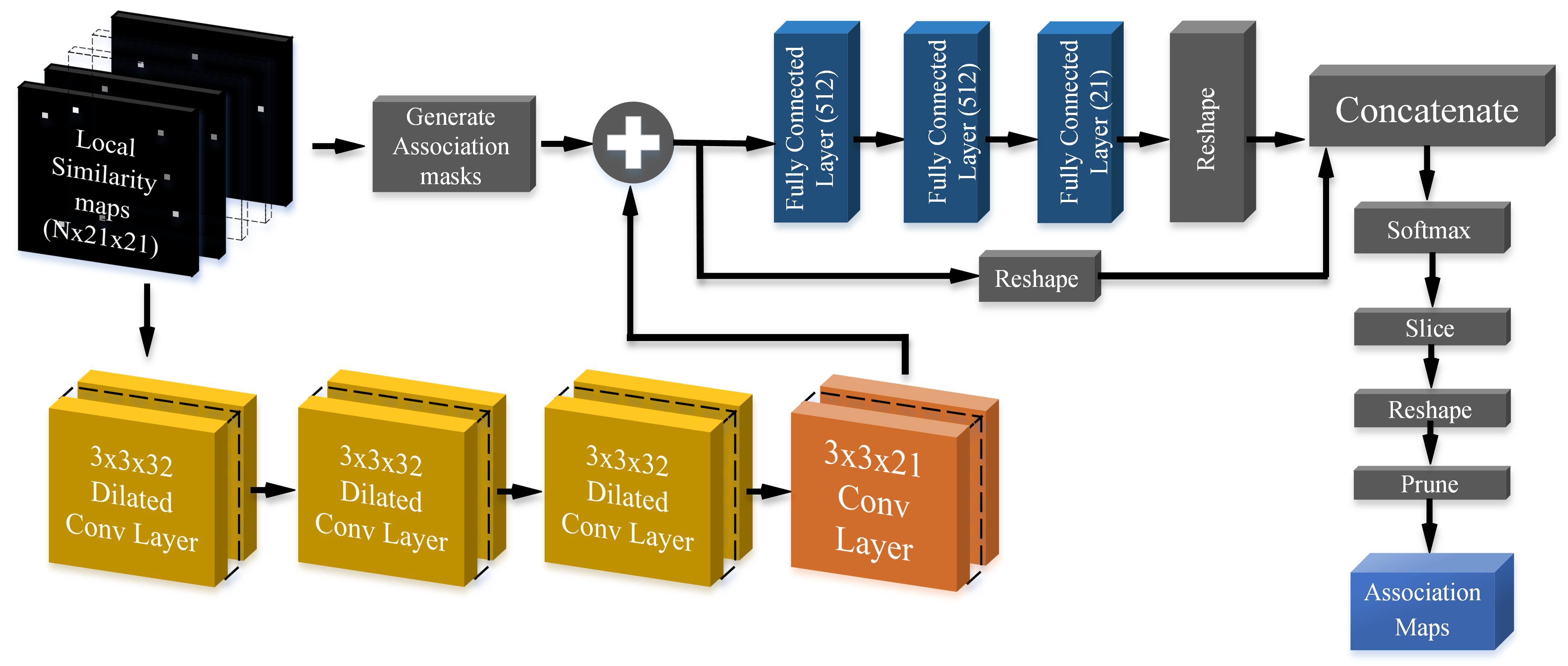}
\end{center}
\caption{The architecture of the proposed association network. The inputs are  local similarity maps from the proposed Siamese network. The outputs are the association maps which provide target-to-detection association and detection probabilities.}
\label{fig_assocnet}
\vspace{-1.5em}
\end{figure*}

\subsection{Data Association Network}
Fig.~\ref{fig_assocnet} introduces our proposed data association network, referred to as \textit{AssocNet}. The purpose of this network is to associate targets to the detections. The \textit{input} to the network is the set of local similarity maps of dimension $N\times21\times21$  obtained from \textit{SimNet}, containing cosine similarity scores for probable target-detection pairs. The \textit{output} from the network is the target-to-detection association probabilities for each existing target.

\par We first describe how our framework handles noisy detections and a varying number of targets. In order to deal with varying number of inputs, we create $N_{max}-N$ extra channels with dummy maps, where $N_{max}$ denotes the maximum number of targets that can be tracked. The dummy maps are a matrix of zeros - a reasonable representation since zero inputs don't have any impact on the output of the convolutional layers. We deal with missed detections by introducing an extra cell for each of the $N_{max}$ targets to account for \textit{spurious detections} and concatenate it to their map of logits as described in the architecture below.

\par \textbf{\textit{Architecture:}} The main building blocks of the \textit{AssocNet} are convolutional, dilated convolutional (d-Conv), and fully-connected layers, with batch-normalization and leaky-ReLU activation used in all the layers. We take advantage of the increased receptive field of dilated convolutions \cite{MSCA_DConv} to compensate for the sparsity of local similarity maps.
\par We now discuss the flow of information through \textit{AssocNet}. The network processes the input using three dilated convolutional (d-Conv) layers with dilation factors of $2, 4$, and $6$ respectively. The neighbouring fields have slightly overlapping fields of view due to increased dilation size \cite{Effective_DConv_Segmentation_RSI}. The convolutional layer enables interactions between these neighbouring units which effectively results in considering all the detections simultaneously while making assignments. Thus to aggregate information, we employ a $3 \times 3$ convolutional layer at the end to compute the maps of logits (the vector of non-normalized predictions).
\par \textit{AssocNet} is to be trained to predict assignment probabilities between a target and its probable detections. Since the locations of probable detections are known in each local similarity map, there is no need to train \textit{AssocNet} to predict assignment probabilities of other locations as zero. To implement this idea, we generate association masks for each local similarity map. In the association masks, cells of probable detections are set to zero, while the other cells are set to a minimum negative number. Then the association masks are added to the map of logits obtained from the convolutional layer with $3\times3\times21$ filters (see Fig.\ref{fig_assocnet}). This maintains the values of the logits computed for probable detections, but makes other logits insignificant for further computation.
\par After masking the maps of logits, \textit{Assocnet} is split into two branches. One branch consisting of fully-connected layers predicts the $N_{max}$ logit values of spurious detections. The other branch reshapes the logit map into 1D vectors to concatenate logits of spurious detections with those of probable detections. This results in a $N_{max}\times(21\times21+1)$ tensor. The softmax then computes the association probabilities for each target, which are is our required output.
\par The association probabilities are sliced and reshaped in order to obtain 2D association maps. The probabilities computed for spurious detections are missed-detection probabilities. Finally, we get rid of the association maps  corresponding to the $N_{max} - N$ dummy channels.

\subsubsection{Loss Function}
Training \textit{AssocNet} can be considered as training a classification problem in which labels are association maps showing the true data association for each existing target. To train the data association network we use a multi-task loss function given by:

\begin{equation}{\label{eq:4}}
L\left({\Theta} \right) = l\left( {\Theta} \right)_{assoc} \ + l\left( {\Theta} \right)_{reg}
\end{equation}

\noindent
where ${\Theta}$ is the set parameters of the association network, $l\left( {\Theta} \right)_{reg}$ is the regularization loss. $l\left( {\Theta} \right)_{assoc}$ is the binary cross-entropy computed for the association maps as follows:

\begin{equation} {\label{eq:5}}
\begin{split}
 q_{vec} &= q_{assoc}^{\left( t \right)} \left( {i,j} \right) \times  \log  \left( {\min \left( {\hat q_{assoc}^{\left( t \right)}\left( {i,j; {\Theta}} \right) + 0.01,1} \right)} \right) 
\\
p_{vec} &= p_{assoc}^{\left( t \right)} \left( {i,j} \right) \times  \log \left( {\min \left( {\hat p_{assoc}^{\left( t \right)} \left( {i,j; {\Theta}} \right)  + 0.01,1} \right)} \right) 
\\ 
 & l\left( {\Theta} \right)_{assoc} =  \sum\nolimits_{t = 1}^{N} \sum\nolimits_{i,j = 1}^{21 + 1} \left( - q_{vec} \right)  + 
  \left(- p_{vec}  \right)
\end{split}
\end{equation}
\noindent
where $ q_{assoc}^{\left( t \right)} \left( {i,j} \right)  = 1 - p_{assoc}^{\left( t \right)}\left( {i,j} \right)$ and $0.01$ is the margin used to ignore negligible  errors  in the predicted probabilities $\hat p_{assoc}^{\left( t \right)}\left( {i,j; {\Theta }} \right)$.

\begin{algorithm}[ht]
\SetAlgoLined
 $\theta_{ex}=0.40$ \tcp*[r]{Existence threshold}
 \While{true}{
   Get Detections $m^{\tau}$ at time $\tau$ \\
 \eIf{$\tau=0$}{
     \ForEach{$m_{i}^0 $}
      {
        Create new track i
      }
   }
   {
       Perform Kalman Filter Prediction;
       
       Predict $P_{e}$ $\forall i \in T$;
       
       $DataAssociation$ for $t^{k}$ and $m^{k}$;
       
       Perform Kalman Filter Update;
       
       Update $P_{e}$ $\forall i \in T$;\\

       $\forall (t_{i}^k,m_{j}^k)$ Update track $i$ with $m_{j}^k$
            
       $\forall (t_{i}^k,None)$ Propagate predicted $t_{i}^k$ to $\tau=k+1$
            
       $\forall (None,m_{j}^k)$ Create a new track; \newline\newline
       $\forall i \in T$ \If{$P_{e_i}^k<\theta_{ex}$}{
       $Prune \quad i$ 
       }
   }
   \caption{Tracker Algorithm}
   \label{alg:tracker_alg}
}
\end{algorithm}
\subsection{Track Management}
The track management module takes care of state estimation, initiation, update, and termination of tracks. We use a Kalman filter for motion prediction and state estimation. We initiate, update and prune tracks with a Bayesian estimation model as specified in \cite{Pak2016JointTD} with a probability of existence $P_{e}$. Our complete tracking algorithm is described in Algorithm \ref{alg:tracker_alg}.

\begin{table*}[ht]
\vspace{1.5em}
\caption{Results on Kitti Test set for 'Car' class}
\centering
\begin{tabular}{c | c | c | c | c | c | c | c | c | c}
\hline \hline
{\bf Method} & {\bf MOTA $\uparrow$} & {\bf MOTP $\uparrow$} & {\bf MT $\uparrow$} & {\bf ML $\downarrow$} & {\bf IDS $\downarrow$} & {\bf FRAG $\downarrow$}\\ 
\hline \hline
MOTBeyondPixels \cite{beyondmot} & {\bf84.24} \% & {\bf85.73} \% & {\bf73.23} \% & {\bf2.77} \% & 468 & 944\\
JCSTD \cite{xiao2010vehicle} & 80.57 \% & 81.81 \% & 56.77 \% & 7.38 \% & {\bf61} & 643\\
3D-CNN/PMBM \cite{scheidegger2018mono} & 80.39 \% & 81.26 \% & 62.77 \% & 6.15 \% & 121 & 613\\
extraCK \cite{gunduz2018lightweight} & 79.99 \% & 82.46 \% & 62.15 \% & 5.54 \% & 343 & 938\\
MDP \cite{xiang2015learning} & 76.59 \% & 82.10 \% & 52.15 \% & 13.38 \% & 130 & {\bf387}\\
\hline
\textit{FANTrack (Ours)} & 77.72 \% & 82.32 \% & 62.61 \% & 8.76 \% & 150 &
812\\
\hline \hline
\end{tabular}
\label{table:testresults}
\vspace{-1.5em}
\end{table*}

\begin{table*}[ht]
\vspace{1.5em}
\caption{Ablation study on KITTI validation set for 'Car' class}
\centering
\begin{tabular}{cccccccc}
\hline \hline
Method & MOTA $\uparrow$ & MOTP $\uparrow$ & MT $\uparrow$ & PT $\uparrow$ & ML $\downarrow$ & IDS $\downarrow$ & FRAG $\downarrow$ \\
\hline \hline
Euclidean+\textit{AssocNet} & 56.16 \% & 84.84 \% & 72.22 \% & 18.51 \% & 9.25 \% & 269 & 320 \\
Manhattan+\textit{AssocNet} & 56.75 \% & 84.83 \% & \bf{73.14} \% & 17.59 \% & 9.25 \% & 265 & 319 \\
Bhattacharyya+\textit{AssocNet} & 56.69 \% & 84.81 \% & 72.22 \% & 18.51 \% & 9.25 \% & 256 & 307 \\
ChiSquare+\textit{AssocNet} & 57.17 \% & 84.81 \% & \bf{73.14} \% & 18.51 \% & \bf{8.33} \% & 262 & 311 \\
\textit{SimNet}+Hungarian & 74.59 \% & \bf{84.92} \% & 65.74 \% & \bf{23.14} \% & 11.11 \% & 26 & 93 \\
\textit{SimNet}+\textit{AssocNet} & \bf{76.52} \% & 84.81 \% & \bf{73.14} \% & 17.59 \% & 9.25 \% & \bf{1} & \bf{54} \\
\hline
\end{tabular}
\label{table:ablation}
\\[5pt]
($\uparrow$ denotes higher values are better. $\downarrow$ denotes lower values are better)
\end{table*}


\section{Experiments}
In this section, we describe the dataset, training parameters, and experimental evaluation results for the tracker built using our proposed data association networks.

\subsection{Dataset}
We used the KITTI Tracking benchmark dataset for training and evaluation of our approach. The KITTI Tracking dataset consists of 21 training sequences and 29 test sequences. As the training sequences have different levels of difficulty, occlusion, and clutter, we split the 20\% of every training sequence for validation. This way, training and validation datasets are not skewed. For training SimNet, we construct a training dataset from the training sequences by generating positive and negative examples in consecutive frames using ground truth information. Geometric transformations (translation, rotation, and scaling) are applied to the ground-truth bounding box parameters to model partial occlusion and detector noise. This gives a large training set in which the ratio of negatives to positives is approximately $18:25$. We trained the object detector using a combined dataset consisting of the KITTI 3D object detection dataset and the 80\% split of the KITTI training dataset mentioned earlier after pre-training on a synthetic dataset \cite{hurl2019precise}.

\subsection{Training Parameters}
\subsubsection{Similarity Network} SimNet is trained with mini-batches of size $128$. Each mini-batch consists of the spatial indices of detections in the global map, the number of targets ($N$), target centroids in $x$-$y$ coordinates, target and detection appearance features, their bounding box parameters, and the labels of each example. To optimize the loss function (\ref{eq1}) we used Adam optimizer and exponentially-decaying learning rate \cite{Adam}. The learning rate is initially set to $1e-5$ and then decreased every $100$ epochs with a base of $0.95$.
\subsubsection{AssocNet} To optimize the loss function in (\ref{eq:4}) we used Adam optimizer and exponentially-decaying learning rate. The learning rate was initially set to $1e-6$ and then decreased every $20$ epochs with a base of $0.95$.

\subsection{Evaluation Metrics}
We use the popular CLEAR MOT metrics \cite{bernardin2008evaluating} for evaluating our tracker. Multiple Object Tracking Accuracy (MOTA) gives us an estimate of the tracker's overall performance. However, this is dependent on the performance of the object detector. Hence, we also look at tracking specific metrics like Mostly Tracked (MT), Mostly Lost (ML), ID Switches (IDS) and fragmentation (FRAG), which evaluate the efficiency of the tracker in assigning the right IDs with reduced switches or fragmentation in the tracks.  

\subsection{Ablation Study}
We do an ablation study to evaluate the components in our approach by comparing them with traditional approaches. Firstly, we study the impact of the similarity network. In Table \ref{table:ablation}, Euclidean and Manhattan denote the baseline distances modeled with the 3D position estimates. Bhattacharyya and ChiSquare metrics are built from the image histograms of the cropped targets and detections to study the image-only configuration. SimNet and AssocNet denote our similarity and Association networks respectively. From Table \ref{table:ablation}, we could infer that conventional similarity approaches were not able to achieve comparable accuracy (MOTA) as the features involved in the computation of the similarity scores were not robust. We also study the impact of our association network by replacing it with a baseline Hungarian approach. Again, we could observe that the baseline approaches like Hungarian couldn't fare better than ours.

\subsection{Qualitative Evaluation}
\begin{figure}
\vspace{1em}
\begin{center}
\includegraphics[width=0.8\linewidth]{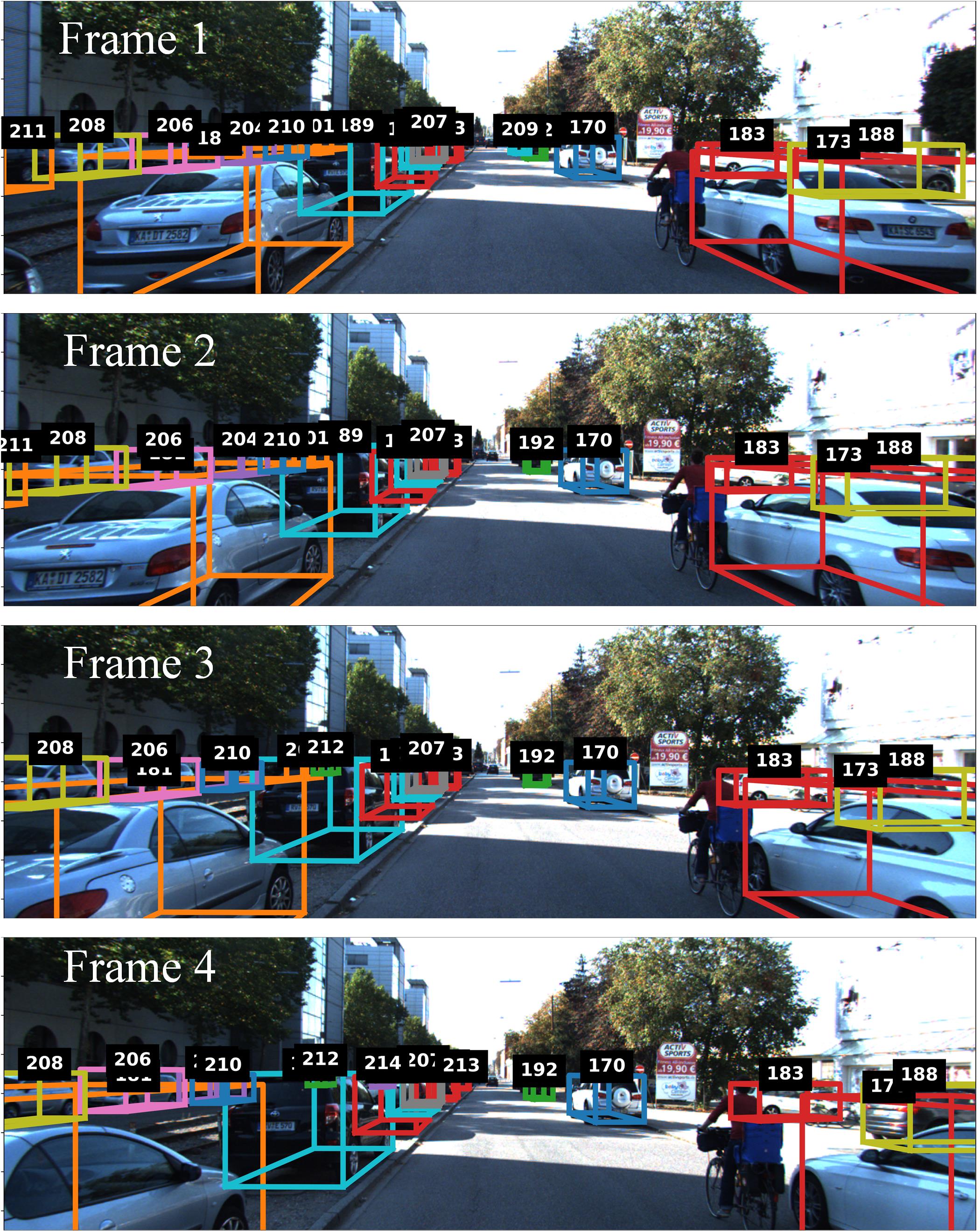}
\end{center}
\caption{Qualitative Evaluation - An example from video 14 in test set where the tracker performs well in a cluttered scene with parked cars.}
\label{fig_qual1}
\vspace{-1em}
\end{figure}

\begin{figure}[hbt]
\vspace{1em}
\begin{center}
\includegraphics[width=0.8\linewidth]{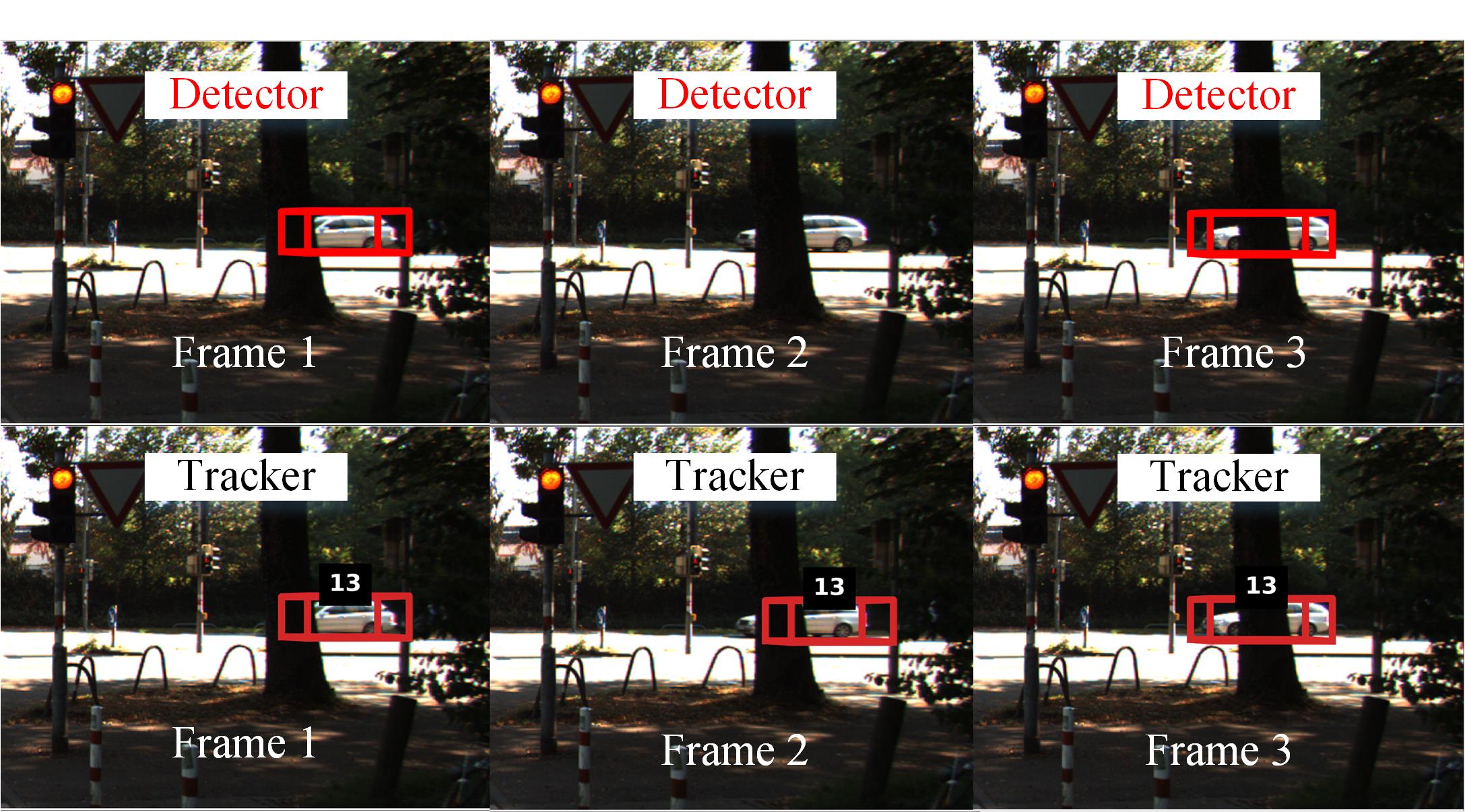}
\end{center}
\caption{Qualitative Evaluation - In this example (video 17 in test set) the detection was missed by the detector and reappears in the next frame. But the tracker was able to successfully maintain the track}
\label{fig_qual2}
\vspace{-1em}
\end{figure}

\begin{figure}[hbt]
\vspace{1.5em}
\begin{center}
\includegraphics[width=0.8\linewidth]{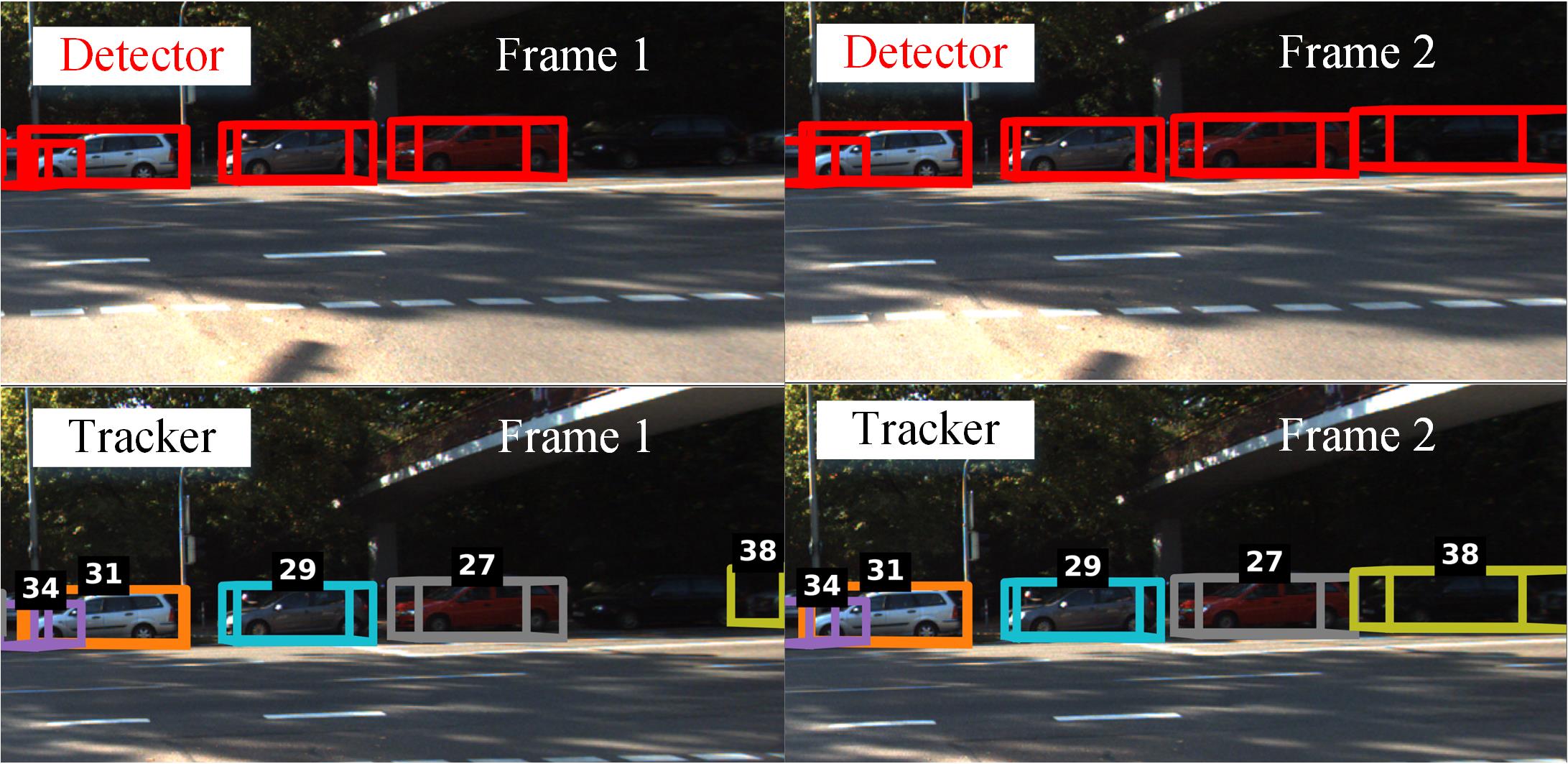}
\end{center}
\caption{Qualitative Evaluation - An example from video 15 in test set where ID switching occurs for Track 38 due to low-lit conditions}
\label{fig_qual3}
\vspace{-1.5em}
\end{figure}

We perform a qualitative evaluation by running our tracker on the KITTI tracking validation and testing sequences. We analyze different scenarios including occlusions, clutter, parked vehicles and false negatives from the detector. Fig. \ref{fig_qual1} shows an example from sequence 0 in the test set. Different tracks representing the vehicles are color coded and the track IDs are displayed for reference. The tracker is able to perform well in spite of the clutter due to the closely parked cars. In Fig. \ref{fig_qual2} we see an example from test sequence 17 in which the false negative by the detector is overcome with the help of the prediction of the tracker. These examples show the robustness of the tracker and its ability to perform better even with an average object detector. There were also some cases where the data association fails and as a result ID switching and fragmentation happen. In  Fig. \ref{fig_qual3} the track 38 was previously assigned to a nearby car but after an occlusion in the detection ID switching happens. This could be due to the low-lit conditions of the two cars.

\subsection{Benchmark Results}
We evaluate our approach on the test sequences on the KITTI evaluation server for the 'Car' class. The results are presented in Table \ref{table:testresults}. Due to the challenging nature of online tracking approach and to do a fair comparison, we only consider published online tracking approaches for our comparison. We achieve competitive results with respect to the state of the art in online tracking with improved MOTP which is better than most of the online methods. Our Mostly Tracked and Mostly Lost (MT \& ML) values are also competitive which show the effectiveness of our data association approach. Further, our approach gives inferences in 3D and KITTI evaluations are done in 2D, which is not completely representative of our approach. It should also be noted that none of these approaches use deep learning for data association. On the other side, we have used a simple Kalman filter for state estimation and motion prediction which could potentially be improved by better tuning of parameters or trying out more sophisticated approaches for track management. 

After optimizing the convolution operation in subsection \textit{A.4} with selective dot products our tracking algorithm has an average runtime of 0.04s per frame (~25 Hz) on Nvidia GeForce GTX 1080 Ti and with a single thread on Intel Core i7-7700 CPU @ 3.60GHz.





\section{Conclusions}
In this paper, we presented a solution to the problem of data association in 3D online multi-object tracking using deep learning with multi-modal data. We have shown that a learning-based data association framework helps in combining different similarity cues in the data and provides more accurate associations than conventional approaches, which helps in increased overall tracking performance. We demonstrated the effectiveness of the tracker built using this model with a multitude of experiments and evaluations and show competitive results in the KITTI tracking benchmark. In the future, we plan to integrate this solution with an object detection framework more tightly and perform end-to-end training. 

\bibliographystyle{IEEEtran}
\bibliography{bibtex/bib/IEEEabrv.bib,bibtex/bib/IEEEexample.bib}{}

\end{document}